\documentclass[10pt,twocolumn,letterpaper]{article}

\usepackage{cvpr}
\usepackage{times}
\usepackage{epsfig}
\usepackage{graphicx}
\usepackage{amsmath}
\usepackage{amssymb}
\usepackage{multirow}
\DeclareGraphicsExtensions{.pdf}

\usepackage[breaklinks=true,bookmarks=false]{hyperref}

\cvprfinalcopy 

\def\cvprPaperID{3594} 

\begin{document}

\title{All You Need is a Few Shifts: Designing Efficient Convolutional Neural Networks for Image Classification}

\author{Weijie Chen, Di Xie, Yuan Zhang, Shiliang Pu\\
Hikvision Research Institute, Hangzhou, China\\
{\tt\small \{chenweijie5, xiedi, zhangyuan, pushiliang\}@hikvision.com}
}

\maketitle

\begin{abstract}

Shift operation is an efficient alternative over depthwise separable convolution. However, it is still bottlenecked by its implementation manner, namely memory movement. To put this direction forward, a new and novel basic component named Sparse Shift Layer (SSL) is introduced in this paper to construct efficient convolutional neural networks. In this family of architectures, the basic block is only composed by 1x1 convolutional layers with only a few shift operations applied to the intermediate feature maps. To make this idea feasible, we introduce shift operation penalty during optimization and further propose a quantization-aware shift learning method to impose the learned displacement more friendly for inference. Extensive ablation studies indicate that only a few shift operations are sufficient to provide spatial information communication. Furthermore, to maximize the role of SSL, we redesign an improved network architecture to Fully Exploit the limited capacity of neural Network (FE-Net). Equipped with SSL, this network can achieve 75.0\% top-1 accuracy on ImageNet with only 563M M-Adds. It surpasses other counterparts constructed by depthwise separable convolution and the networks searched by NAS in terms of accuracy and practical speed. 

\end{abstract}

\section{Introduction}

\begin{figure}[tp]
\begin{center}
\includegraphics[scale=0.35]{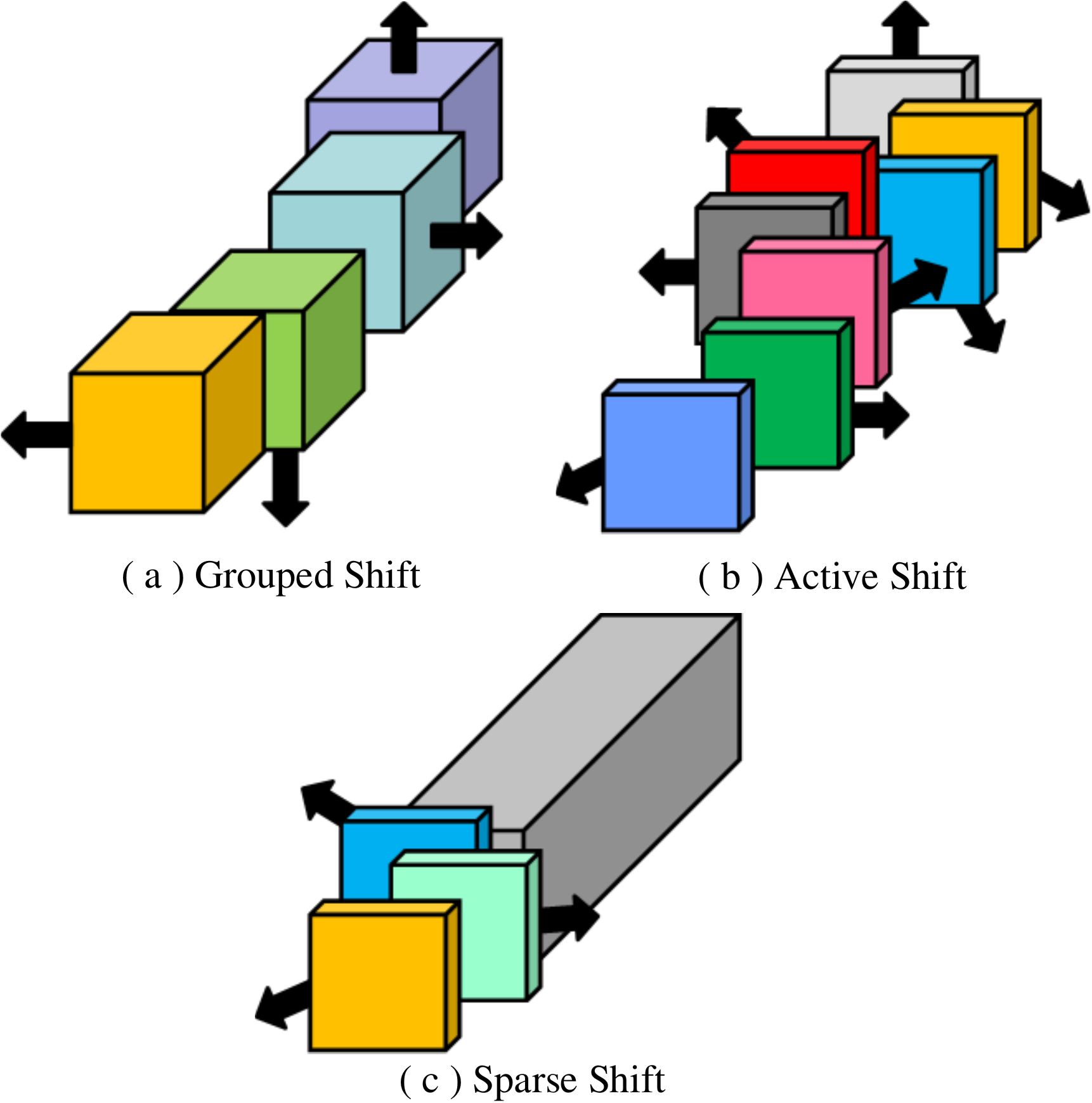}
\end{center}
\caption{The comparison of different shift operations applied to feature maps.}
\label{shiftCompare}
\end{figure}

Owing to the amazing performance of convolutional neural networks (CNNs), it becomes a big trend to apply CNNs to practical application scenarios. However, it is hindered by their substantial computational cost and storage overhead, which motivates lots of researchers and engineers to gush into this subject.


One of the useful solutions to tackle this problem is to design accurate and compact neural network architectures directly. A well-designed network topology as well as a hardware-friendly basic component can bring about surprising breakthroughs.  Recently, a popular basic component named depthwise separable convolution is welcomed to design lightweight architectures, such as MobileNet \cite{MobileNets} and ShuffleNet \cite{ShuffleNets}. Despite its lower float point operations (FLOPs), it is inefficient to implement in practice because of the fragmented memory footprints. To jump out of the constraint of depthwise separable convolution, ShiftNet \cite{wu2018shift:} propose another alternative, say shift operation, to construct architectures cooperated with point-wise convolution. In this network, shift operation provides spatial information communication through shifting feature maps, which makes the followed point-wise convolution layer not only available for channel information aggregation but also available for spatial information aggregation.

\begin{figure*}[tp]
\begin{center}
\includegraphics[scale=0.19]{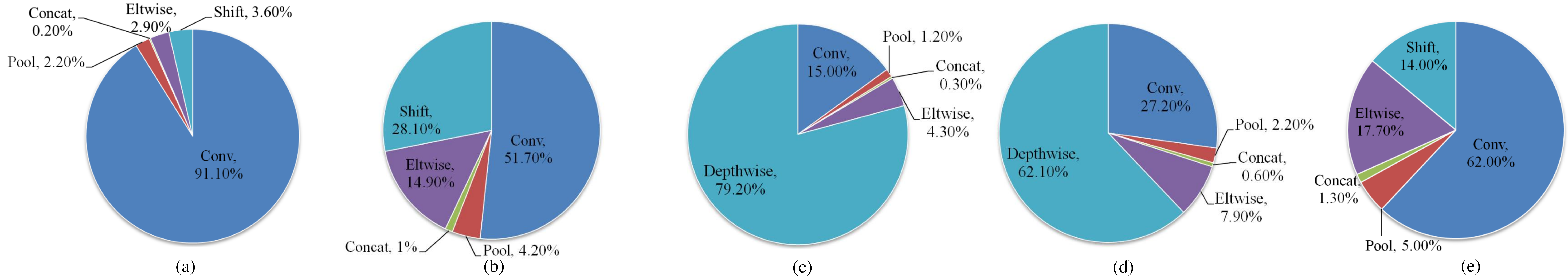}
\end{center}
\caption{The practical runtime analysis. For clear comparison, both batch-normalization and ReLU layers are neglected since they can be merged into convolutional layer for inference. Also data feeding and preprocessing time are not considered here. Results are achieved under Caffe with mini-batch 32. They are averaged from 100 runs. (a) ShiftNet-A \cite{wu2018shift:} on CPU (Intel Xeon E5-2650, atlas). (b) ShiftNet-A on GPU (TITAN X Pascal, CUDA8 and cuDNN5). (c) Shift layers in ShiftNet-A are replaced by depthwise separable convolution layers. (d) Depthwise separable convolution layers with kernel size 5 are replaced by the ones with kernel size 3. (e) ShiftNet-A with 80\% shift sparsity on GPU (Shift sparsity denotes the ratio of unshifted feature maps).}
\label{shiftTime}
\end{figure*}
 
 In order to compare these two basic components, we decompose the occupied time of each basic component of ShiftNet for detailed analysis on both compute-bound and memory-bound computation platforms. As illustrated in Fig.\ref{shiftTime} (a) and (b), shift operation occupies 3.6\% of runtime on CPU, but occupies 28.1\% on GPU, indicating that shift operation still occupies considerable runtime on memory-bound computation platforms due to memory movement. As for depthwise separable convolution, in MobileNetV2, it occupies about 36\% runtime on GPU. However, it is unfair to compare these two components in two different architectures. For fair comparison, we use the same architecture with ShiftNet and only replace shift operation by depthwise separable convolution to test its inference time. As shown in Fig.\ref{shiftTime} (c), it occupies 79.2\% of runtime on GPU which seriously mismatches its theoretical FLOPs. From this viewpoint, shift operation is significantly superior to depthwise separable convolution. Also, another attractive characteristic of shift operation is its irrelevance of computational cost to kernel size, while the practical runtime of depthwise separable convolution is strongly influenced by kernel size. As illustrated in Fig.\ref{shiftTime} (c) and (d), the occupied runtime of depthwise separable convolution is lowered to 62.1\% after decreasing the kernel size 5\footnote{In ShiftNet, there are 11 shift layers with kernel size 5.} to 3.

Despite the superiority of shift operation in terms of practical runtime to depthwise separable convolution, it is still bottlenecked by its implementation, namely memory movement. Here naturally comes a question, \emph{is each shift operation really necessary?} Those moving memory can be reduced if the meaningless shifts are eliminated. Bringing this question, we make a further study about shift operation. To suppress redundant shift operation, penalty is added during its optimization. We surprisingly find that a few shift operations are actually sufficient to provide spatial information communication. It can provide comparable performance with shifting a small portion of feature maps. We name this type of shift layer as \emph{Sparse Shift Layer} (SSL) in order to distinguish from other types of shift layers as shown in Fig.\ref{shiftCompare}. As shown in Fig.\ref{shiftTime} (e), it can significantly reduce the occupied time of shift operation after inducing sparsity.

The prerequisite of SSL is to ensure shift operation learnable. A common solution is to relax the displacement from integer to real-value and relax shift operation to bilinear interpolation so as to make it differentiable \cite{activeShift}. However, interpolation cannot bring the same inference benefit as shift operation. Borrowing the idea from QNN \cite{hubara2016quantized}, we propose a quantization-aware shift learning method to enable shift operation differentiable while avoiding interpolation during inference.

When designing the compact network architecture, a straightforward guideline is to ensure the information flow while maintaining the feature maps diversity. We hope it can contain label-related information as abundant as possible in the limited feature space. However, the feature maps usually tend to collapse into a small subset, which does not fully exploit the limited feature space. To ease this problem, we design a novel network architecture FE-Net as shown in Fig.\ref{architecture}, which involves feature maps into computation progressively as layer increases to impose diversity while avoiding redundant overhead. 

In this paper, we mainly conduct experiments on image classification benchmarks. Extensive ablation studies on CIFAR-10 and CIFAR100 validate the impact of SSL. Furthermore, we carry out experiments on a large-scale image classification dataset ImageNet to confirm the efficiency and the generalization of SSL. With network architecture improvement, we surpass ShiftNet and AS-ResNet \cite{activeShift} by a large margin. \emph{It is worth highlighting that our network even surpasses other counterparts composed by depthwise separable convolution.} We achieve {\bfseries 75.0\%} top-1 accuracy on ImageNet with {\bfseries 563M} M-Adds. This is the first time the compact networks can achieve such high accuracy in this level of computational cost without using depthwise separable convolution. Equipped with \emph{Squeeze-and-Excitation} module \cite{SE} in a proper way, our network can be further boosted to {\bfseries 76.5\%} top-1 accuracy with {\bfseries 566M} M-Adds.

In summary, our main contributions are listed as follows:
\begin{itemize}
\item A new basic component named \emph{Sparse Shift Layer} is introduced to build fast and accurate neural networks, which can eliminate meaningless memory movement. Beyond this, through extensive ablation studies, we find that only a few shift operations are sufficient to provide spatial information communication, which will inspire more exploration in the development of compact neural networks.
\item A quantization-aware shift learning method is proposed to ensure shift operation learnable while avoiding interpolation during inference.
\item An improved compact network architecture is designed to fully exploit the capacity of the limited feature space. Combining it with SSL, we achieve state-of-the-art results in classification benchmarks in terms of both accuracy and inference speed.
\end{itemize}

\section{Related Works}

Over the past few years, more and more approaches are proposed to lighten neural networks in terms of storage, computation and the practical inference time, while keeping their performance powerful. We divide these related methods into the following two parts from the view of whether a pretrained model is given.

\subsection{Neural Networks Compression}

To compress a given pretrained model into a lightweight one,  there exist four different approaches: 1) Pruning \cite{Hassibi1993Second,lecun1990optimal,Han2015Learning,Wen2016Learning,Li2016Pruning,Molchanov2016Pruning,He2017Channel,Luo2017ThiNet,slimming,Net-Trim,LDRF} aims to remove unimportant parameters and turns weight matrices into sparse ones. 2) Tensor decomposition \cite{Denton2014Exploiting,jaderberg2014speeding,lebedev2015speeding-up,kim2016compression,Wen2017Coordinating,Tai2015Convolutional} exploits the channel or spatial redundancy of weight matrices and seeks their low-rank approximations. 3) Quantization \cite{gong2014compressing,hubara2016quantized,rastegari2016xnor-net:} adopts low bits instead of float point representation for each weight parameter. 4) Knowledge distillation \cite{Hinton2015Distilling,sau2016deep} transfers the knowledge from teacher models to lightweight student models.

These methods can effectively compress neural networks into small ones. However, their performances heavily depend on the given pretrained model. Without architecture improvement, the accuracy cannot go a step further.

\subsection{Compact Networks Development} 

How to design a compact neural architecture is a popular research topic recently. Some related works \cite{deeproots,igcv} used group convolution to construct compact networks. The most famous work, MobileNet \cite{MobileNets}, adopts depthwise separable convolution to build an accurate and lightweight network, which moves forward a big step in this field. After that, lots of researchers follow these works and design more compact and powerful architectures, such as ShuffleNet, MobileNetV2, ShuffleNetV2, IGCV2 and so on \cite{ShuffleNets,sandler2018mobilenetv2:,shufflenetv2,igcv2}. However, even though depthwise separable convolution only needs little theoretical computation cost, it is difficult to implement efficiently in practice since the arithmetic intensity is too low.

\cite{wu2018shift:} provide an alternative named shift operation which only shifts feature maps without computation. A compact network can be constructed by interleaving this operation with point-wise convolutions. \cite{activeShift} propose a method to make shift operation learnable, which means the receptive field of each layer can be learnt automatically. The existing problem is that this operation still occupies considerable inference time because it is implemented by memory movement. This is exactly what we want to solve in this paper.

\section{Background}
We first review the standard shift operation, which can be formulated as follows:
\begin{equation}\label{1}
O_{c,i,j}=I_{c,i+\alpha_{c},j+\beta_{c}}
\end{equation}
where $I$ and $O$ are the input and output feature maps, respectively. $c$ is the channel index. $i$ and $j$ denote the spatial position. $\alpha_{c}$ and $\beta_{c}$ denote the horizontal and vertical displacement assigned to $c_{th}$ input feature map. The parameter number of $\alpha$ and $\beta$ is separately equivalent to the channel number of input feature maps, which is almost negligible compared with the parameters of convolution layer. 

{\bfseries Grouped shift.} In the work of \cite{wu2018shift:}, for a shift operation with kernel size $K$, the input feature maps are evenly divided into $K^{2}$ groups, and each group is assigned one displacement as illustrated in Fig.\ref{shiftCompare}(a). This displacement assignment can be formulated as follows:
\begin{equation}\label{2}
\begin{array}{lr}
\alpha_{c}=\lfloor \lfloor c/K^{2} \rfloor \,\,\,/\,\,\,K\rfloor - \lfloor K/2 \rfloor,&\\
\beta_{c}=\lfloor c/K^{2} \rfloor\bmod K - \lfloor K/2 \rfloor,&\\
\end{array}
\end{equation}
where $\lfloor \cdot \rfloor$ denotes floor function. However, the heuristic assignment is not task-driven. The kernel size of each shift operation is set through lots of trial-and-error experiments, and the uniform distribution of the displacement is not generally suitable for every task.

{\bfseries Active shift.} To solve this problem, \cite{activeShift} proposes a method to make $\alpha$ and $\beta$ differentiable, which relaxes the integer constraint of $\alpha$ and $\beta$ to real value and relaxes shift operation to bilinear interpolation. In this manner, Eqn.\ref{1} can be relaxed as follows:
\begin{equation}\label{3}
O_{c,i,j}=\!\!\!\!\!\!\sum_{(n,m)\in\Omega}\!\!\!\!\!\!\!I_{c,n,m}\cdot(1\!-\!\left| i+\alpha_{c}-n\right|)(1\!-\!\left| j+\beta_{c}-m\right|)
\end{equation}
where $\Omega$ is the neighbor set of $(i+\alpha_{c},j+\beta_{c})$ composed by four nearest integer points. Hence, $\alpha$ and $\beta$ can be optimized adaptively by gradient descent optimizers through backpropagation. This shift pattern is illustrated in Fig.\ref{shiftCompare}(b).
\section{Designing Efficient Convolutional Neural Networks with a Few Shifts}

It demonstrates in the works of \cite{activeShift,wu2018shift:} that shift operation can provide receptive field for spatial information communication in ConvNets. However, not each feature map is required to shift. Redundant shift operation will bring redundant memory movement and further impact the inference time of neural network. Starting from this point, we develop a method in this section to build efficient ConvNets with fewer shift operations.

\subsection{Sparsifying Shift Operation}
To avoid meaningless memory movement, we add displacement penalty to eliminate useless shift operation in loss function. Also, it can avoid diffusion of shift learning since a big displacement will induce useful boundary information loss especially for those feature maps with lower resolution. To this end, we add L1 regularization to $\alpha$ and $\beta$ to penalize redundant shifts, which is formulated as follows:
\begin{equation}\label{4}
\begin{array}{clr}
\mathcal{L}_{total}\!\!=\!\!\sum_{(x,y)} \mathcal{L}(f(x \mid W, \alpha, \beta), y) + \lambda \mathcal{R}(\alpha,\beta) &\\
\mathcal{R}(\alpha, \beta)=\parallel \alpha \parallel_{1}+\parallel \beta \parallel_{1} &\\

\end{array}
\end{equation} 
where $(x,y)$ is the input data and its corresponding label, $W$ denotes the trainable parameters except $\alpha$ and $\beta$, $f(\cdot)$ outputs the predicted label, $\mathcal{L}(\cdot)$ is the loss function of neural networks, and $\lambda$ balances these two terms. 

With such sparsity-induced penalty, we can adopt minimum memory movement to build an accurate and fast neural network. We name this new component as sparse shift layer
(SSL), which is illustrated in Fig.\ref{shiftCompare}(c), to distinguish from the previous shift operations .

\subsection{Quantization-aware Shift Learning}

Despite flexibility and sparsity are introduced, some problems remain unsolved. Although the integer constraint of $\alpha$ and $\beta$ is relaxed to real value for the sake of learning shift operation, it weakens the inference advantage of shift operation to some extent since interpolation still needs multiplications while standard shift operation only needs memory movement during inference.

Inspired by the method of training quantization neural networks \cite{hubara2016quantized}, we propose a quantization-aware shift learning approach to make these problems tractable. In this approach, we aim to quantize the displacement back to integer during feed-forward, while keeping shift operation still learnable.

{\bfseries Feed-forward.} We use integer approximation of $\alpha$ and $\beta$ to recover shift operation instead of interpolation, which can be formulated as follows:
\begin{equation}\label{6}
O_{c,i,j}=I_{c,i+\left|\alpha_{c}\right|_{\dag},j+\left|\beta_{c}\right|_{\dag}}
\end{equation}
where $\left| \cdot \right|_{\dag}$ denotes the rounding approximation of real value. In this way Eqn.\ref{3} is actually converted back to Eqn.\ref{1} through quantization, meaning that we apply shift operation instead of interpolation to compute the loss of network.

{\bfseries Back-propagation.} Different from feed-forward phase, real-valued shift is required to compute their gradients and optimized through Stochastic Gradient Descent (SGD). According to Eqn.\ref{3}, the gradients of loss with respect to $\alpha$ and $\beta$ are formulated as follows:
\begin{equation}\label{7}
\begin{array}{lr}
\begin{aligned}
\frac{\partial\mathcal{L}}{\partial\alpha_{c}}\!\!=&\sum_{i}^{w}\sum_{j}^{h}\frac{\partial\mathcal{L}}{\partial O_{c,i,j}}\sum_{(n,m)\in\Omega}I_{c,n,m}\cdot\\
&(1-\left| j+\beta_{c}-m\right|)\cdot \mathrm{Sign}(n-i-\alpha_{c})&\\
\frac{\partial\mathcal{L}}{\partial\beta_{c}}\!\!=&\sum_{i}^{w}\sum_{j}^{h}\frac{\partial\mathcal{L}}{\partial O_{c,i,j}}\sum_{(n,m)\in\Omega}I_{c,n,m}\cdot\\
&(1-\left| i+\alpha_{c}-n\right|)\cdot \mathrm{Sign}(m-j-\beta_{c})
\end{aligned}
\end{array}
\end{equation}
where $w$ and $h$ are the spatial size of input feature maps. And $\mathrm{Sign(\cdot)}$ is a function to output $+1$ or $-1$ according to the sign of input value.

As to back-propagate the gradients of loss with respect to the feature maps from higher layers to shallow layers, both Eqn.\ref{3} and \ref{6} work to compute the partial derivation. Consider Eqn.\ref{6} is the actual feed-forward process we apply instead of Eqn.\ref{3}. It is more reasonable and efficient to adopt Eqn.\ref{6} to compute the gradients, which is formulated as:
\begin{equation}\label{8}
\frac{\partial\mathcal{L}}{\partial I_{c,i,j}}=\frac{\partial\mathcal{L}}{\partial O_{c,i-\left|\alpha_{c}\right|_{\dag},j-\left|\beta_{c}\right|_{\dag}}}
\end{equation}
This is an inverse memory movement compared with Eqn.\ref{6}.

{\bfseries Discussion.} After training, the rounding approximation of displacement is preserved and only shift operation is executed during inference. What's more, another surprising by-product is that this method turns L1 regularizer into a \emph{truncated} regularizer, which will shrinks more small displacement towards exact zero.

\subsection{Network Architecture Improvement} 

\begin{figure}[tp]
\begin{center}
\includegraphics[scale=0.4]{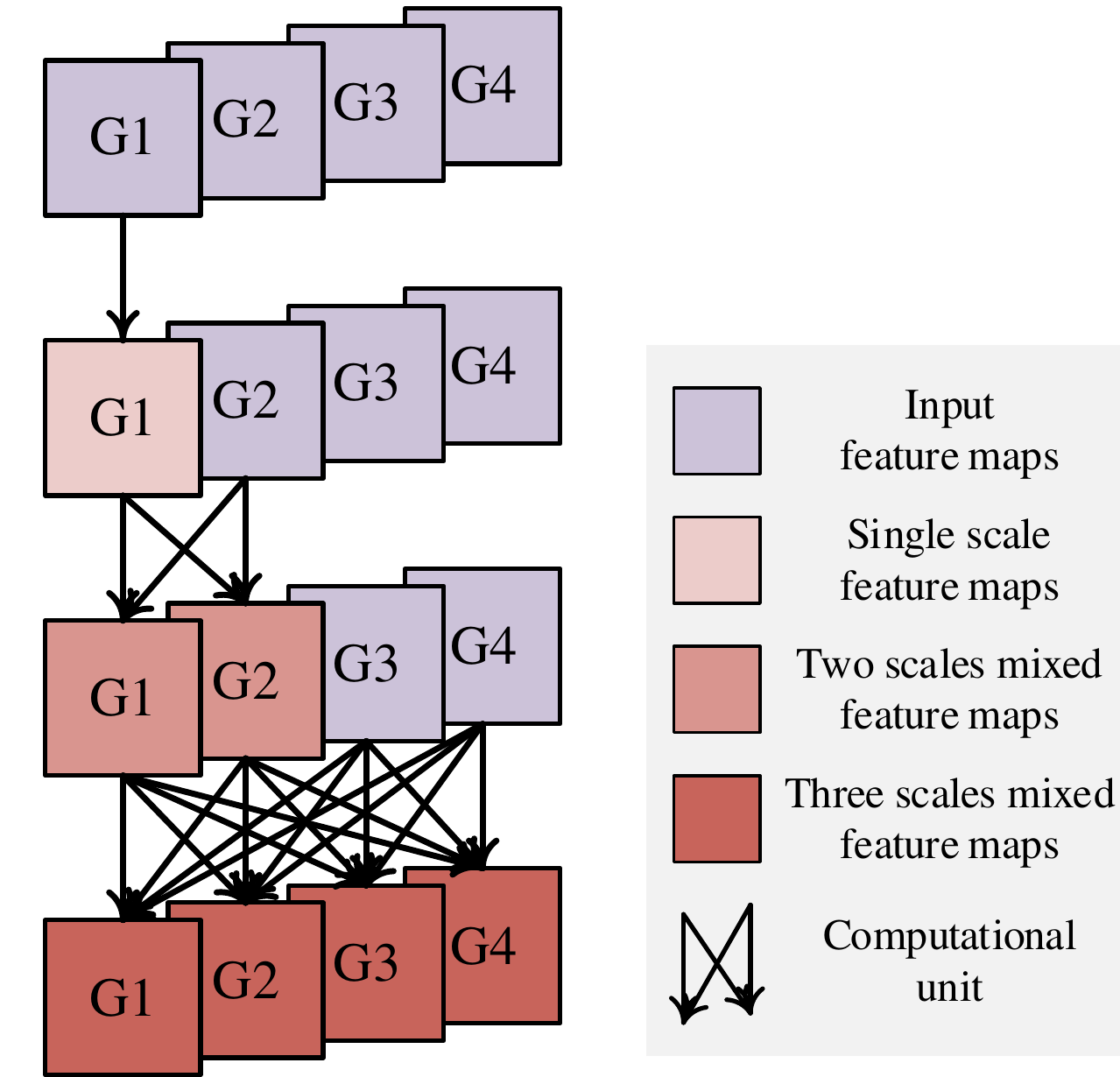}
\end{center}
\caption{Fully-Exploited computational Block (FE-Block). Only a subset of feature maps is involved into optimization at each basic computational unit. For each resolution, the feature maps are progressively mixed in as layer increases. For a computational block with $n=3$ basic units as shown in this figure, we evenly divide the input feature maps into $2^{n-1}$ parts, and involve $\frac{2^{l-1}}{2^{n-1}} (l=1,\ldots,n)$ feature maps into optimization each layer. In this paper, the computational unit is implemented as \emph{inverted bottlenecks} \cite{sandler2018mobilenetv2:}.}
\label{architecture}
\end{figure}

The network capacity is not always fully exploited. As demonstrated in the works of cross-channel decomposition \cite{zhang2016accelerating}, feature maps usually tend to collapse into a small subset. From this perspective, not each feature map is necessary to be involved into the next layer's convolution. According to this insight, we redesign an improved network architecture to ease this problem.


{\bfseries Network architecture.} In this section, we propose a Fully-Exploited Network (FE-Net) composed by the block as shown in Fig.\ref{architecture}. In this block, only a subset of feature maps are involved into computation, and the remaining ones directly propagate to the next layer to ensure information flowing which can be formulated as follows: 
\begin{equation}\label{9}
\begin{array}{lr}
I_{1}, I_{2}\Leftarrow I&\\
O = f(I_{1}) \parallel I_{2}
\end{array}
\end{equation} 
where $I$ and $O$ mean the input and output feature maps. $\Leftarrow$ denotes channel-wise split and $\parallel$ denotes channel-wise concatenation. In practice, this computational pattern can be implemented for efficiency as that $I_{1}$ joins into computation and its output $f(I_{1})$ is rewritten back to the original memory position of $I_{1}$. The remaining feature maps $I_{2}$ do not require any operations. As layer increases, we mix more feature maps into computation. In this way, each input feature map is involved into optimization at last, and multi-scale feature maps are obtained for prediction. We empirically prove its validation in section \ref{ImageNet_Experiments}.

\begin{figure}[tp]
\begin{center}
\includegraphics[scale=0.25]{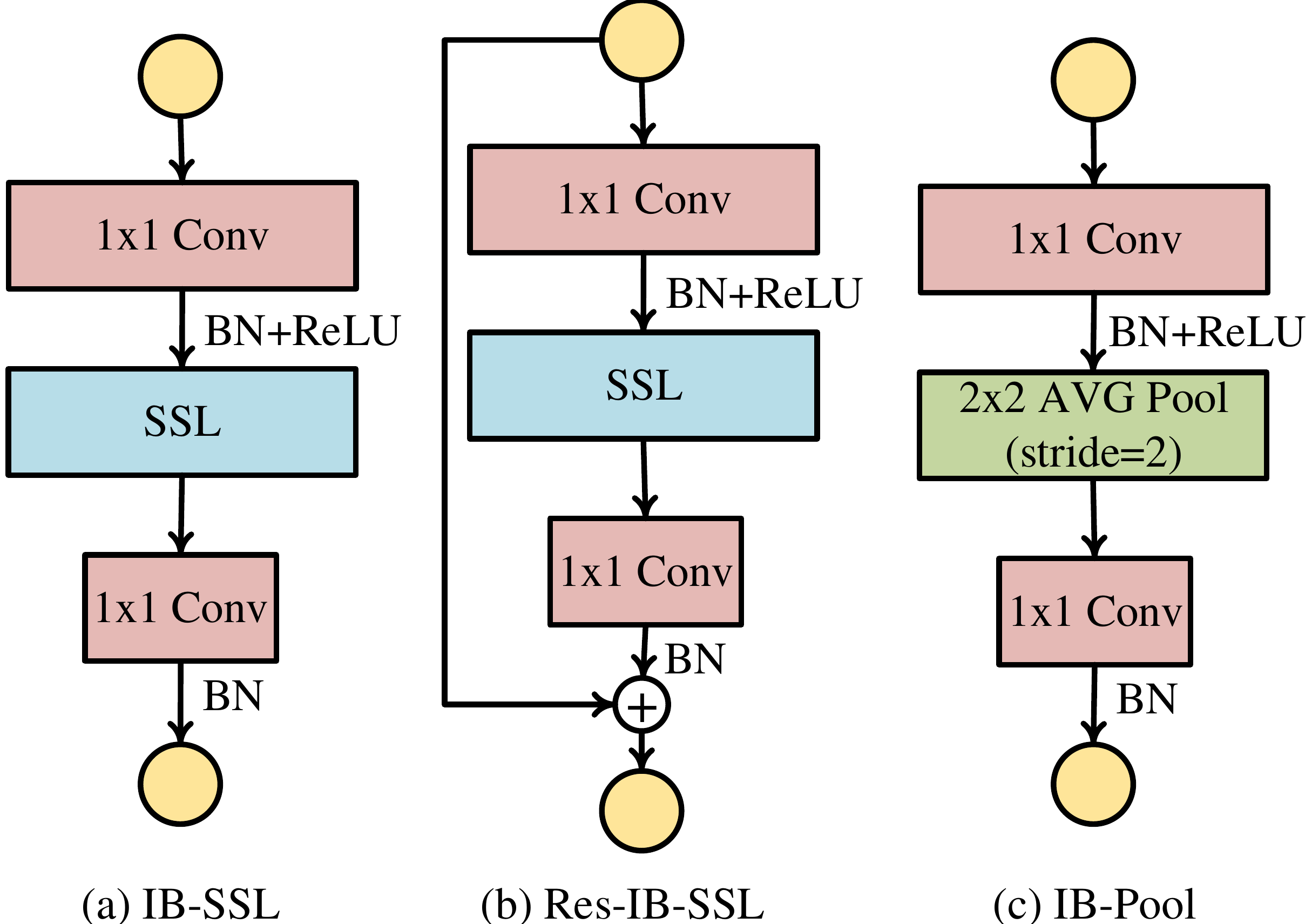}
\end{center}
\caption{The basic computational units for FE-Block. (a): the basic unit without skipping connection; (b): the basic unit with skipping connection; (c): the basic unit for spatial down sampling (2$\times$). Note that in (a) and (b), it is SSL that provides receptive fields. (IB is short for inverted bottleneck.)}
\label{computationalCell}
\end{figure}

{\bfseries Basic computational unit.} In this paper, we adopt inverted bottlenecks \cite{sandler2018mobilenetv2:} as basic computational units to build our efficient networks as shown in Fig.\ref{computationalCell}. Without any specific statement, their expansion rate is always set to 6 by default, which means the first 1$\times$1 Conv is always used to expand the input channel by 6 times. To combine the advantages of \emph{residual learning} \cite{He2016Deep}, for each computation block as shown in Fig.\ref{architecture}, we mainly adopt Fig.\ref{computationalCell}(b) as the basic computational unit except the last one. For the last computational unit at each computation block, we use Fig.\ref{computationalCell}(a) to change the channel number for the next computational block, or use Fig.\ref{computationalCell}(c) for spatial down sampling.

\section{Experiments}\label{Experiments}
In this section, we first carry out several ablation experiments on CIFAR10 and CIFAR100  \cite{cifar} to demonstrate the effect of SSL. In these experiments, we prove that it is enough to provide spatial information communication and build compact ConvNets with only a few shift operations. Then we conduct experiments on ILSVRC-2012 \cite{Russakovsky2015ImageNet} to assess its generalization ability to a large-scale dataset.
\subsection{Benchmarks and Training Settings}

\begin{table*}[tp]
\newcommand{\tabincell}[2]{\begin{tabular}{@{}#1@{}}#2\end{tabular}}
\begin{center}
\begin{tabular}{|c|c|c|c|c|c|}
\hline
depth&Networks&$\lambda$&\tabincell{c}{Accuracy\\CIFAR10 / CIFAR100}&Params / FLOPs&\tabincell{c}{Shift Sparsity\\CIFAR10 / CIFAR100}\\
\hline
\multirow{7}{*}{20}&ResNet \cite{wu2018shift:}&-&91.4\% / 66.3\%&0.27M / 81M&-\\
\cline{2-6}
&ShiftResNet (GroupedShift) \cite{wu2018shift:}&-&90.6\% / 68.6\%&\multirow{6}{*}{0.16M / 53M}&11.1\%\\
\cline{2-4}\cline{6-6}
&\multirow{4}{*}{ShiftResNet (SSL)} & 0 &91.7\% / 69.2\%&&12.1\% / 10.3\%\\
\cline{3-4}\cline{6-6}
&&1e-4&91.1\% / 69.2\%&&66.6\% / 41.2\%\\
\cline{3-4}\cline{6-6}
&&4e-4&90.4\% / 67.7\%&&91.7\% / 80.0\%\\
\cline{3-4}\cline{6-6}
&&5e-4 &89.8\% / 67.7\%&&93.5\% / 86.1\%\\
\cline{2-4}\cline{6-6}
&ShiftResNet (1x1 only)&-&81.5\% / 56.7\%&&100\%\\
\hline
\multirow{7}{*}{56}&ResNet \cite{wu2018shift:}&-&92.0\% / 69.3\%&0.86M / 251M&-\\
\cline{2-6}
&ShiftResNet (GroupedShift) \cite{wu2018shift:}&-&92.7\% / 72.1\%&\multirow{6}{*}{0.55M / 166M}&11.1\%\\
\cline{2-4}\cline{6-6}
&\multirow{4}{*}{ShiftResNet (SSL)} & 0 &93.8\% / 72.4\%&&12.8\% / 11.4\%\\
\cline{3-4}\cline{6-6}
&&1e-4&92.9\% / 71.7\%&&87.8\% / 73.8\%\\
\cline{3-4}\cline{6-6}
&&4e-4&91.9\% / 71.1\%&&97.4\% / 94.6\%\\
\cline{3-4}\cline{6-6}
&&5e-4  &91.8\% / 69.9\%&&98.0\% / 96.1\%\\
\cline{2-4}\cline{6-6}
&ShiftResNet (1x1 only)&-&82.5\% / 56.1\%&&100\%\\
\hline
\end{tabular}
\end{center}
\caption{The analysis of SSL on CIFAR10 and CIFAR100}
\label{cifar_result}
\end{table*}

{\bfseries CIFAR10 / CIFAR100} \cite{cifar} are the datasets for 10-categories and 100-categeories image classification, respectively. Both of them consist of 50k images for training and 10k images for testing with resolution 32 $\times $ 32. 

In the experiments on CIFAR, we choose ShiftResNet \cite{wu2018shift:} which is built by \emph{CSC} modules to evaluate the ability of SSL. Note that a \emph{CSC} module is composed by a shift layer sandwiched between a 1$\times$1 Conv layer for dimension ascending and a 1$\times$1 Conv layer for dimension descending. Only the shift layer in this module is leveraged for spatial information communication. Replacing shift layer by SSL, through adjusting hyperparameter $\lambda$ in Eqn.\ref{4}, we study how many shift operations are required at least to maintain the performance of ShiftResNet.

We use ShiftResNet-20 and ShiftResNet-56 with expansion rate 6 as two representatives for ablation study. We train these networks by two GPUs with mini-batch 128 and base learning rate 0.1. As the same with \cite{wu2018shift:}, the learning rate decays by a factor of 10 after 32k and 48k iterations, and the training stops after 64k iterations. Specifically, we stop the training of SSL after 48k iterations in order to fix the learned shift pattern. For data augmentation, only horizontal flipping and random cropping are adopted. We use L2 regularization to shift values in the following experiments since we find that the result of L2 regularization is slightly better than L1.

{\bfseries ImageNet2012} \cite{Russakovsky2015ImageNet} is a large-scale image classification benchmark with 1.28 million images for training and 50k images for validation. As is known, it is challenging to perform well on such large-scale dataset with lightweight neural networks. In order to boost the performance of the networks built with SSL by a further step, we redesign the neural network architecture as shown in Fig.\ref{architecture} to fully exploit the limited network capacity.

In the experiments on ImageNet, we use SGD to train the networks with mini-batch 1024, weight decay 0.00004 and momentum 0.9. Training is started by a learning rate 0.6 with linear decaying policy and is stopped after 480 epochs, while the training of SSL is stopped after 240 epochs. The entire training iteration is comparable with \cite{igcv3,PNASNET,sandler2018mobilenetv2:,shufflenetv2}. For data augmentation, we scale the short-side of images to 256 and adopt $224\times 224$ random crop as well as horizontal flip to augment the training dataset. Also, to further rich the training images, more image of distortions are provided as used in Inception training \cite{rethinking,MobileNets}. But it will be withdrawn in last several epochs.  At the validation phase, we only center crop the feeding resized images to $224\times 224$ and present the results with single-view approach. 

\subsection{Ablation Study}

We explore the characteristic of SSL from three terms: (i) grouped shift vs. sparse shift; (ii) deep networks vs. shallow networks; (iii) the settings of $\lambda$. 

{\bfseries Grouped shift vs. sparse shift.} As shown in Tab.\ref{cifar_result}, without shift penalty, the results of shift learning are superior to that of heuristic setting on both CIFAR10 and CIFAR100. Through shift learning, the network can adaptively adjust the displacement and direction of shift operation according to different tasks and different datasets. With shift penalty, it can eliminate a great portion of shift operations while keeping the accuracy of the network comparable with original network. Even with more than 90\% sparsity to shift operation, the network can maintain a quite good performance, which suggests that only a few shift operations play crucial roles on communicating spatial information for image classification.

{\bfseries Deep networks vs. shallow networks.} We analyze the sparsity of SSL on CIFAR10 / CIFAR100 with a shallow network and a deeper one, say ShiftResNet-20 and ShiftResNet-56. As shown in Tab.\ref{cifar_result}, the shift sparsity on ShiftResNet-56 is more than ShiftResNet-20. It can provide good performance on CIFAR10 / CIFAR100 with even over 95\% sparsity on ShiftResNet-56. Increasing depth brings more redundancy in the shift layer.


{\bfseries Different settings of $\lambda$.} We increase $\lambda$ from 0 to 5e-4 and find that a majority of shift operation is eliminated progressively while the accuracy of the networks decline a little. Here SSL ($\lambda$=0) is actually equivalent to quantization-aware Active Shift. When we increase $\lambda$ significantly, we shrink all displacement to zero, which means the basic modules are all composed by 1$\times$1 Conv layers and only three pooling layers in the network provide for spatial information communication. In this case, the accuracy drops a lot, which reflects from another side that such a few shifts really matter a lot for spatial information communication. Let us take ShiftResNet-56 on CIFAR100 as an example. Its accuracy can be boosted from 56.1\% to 69.9\% with only 3.9\% feature maps shifted.

\subsection{Case Study}

\begin{figure}[tp]
\begin{center}
\includegraphics[scale=0.2]{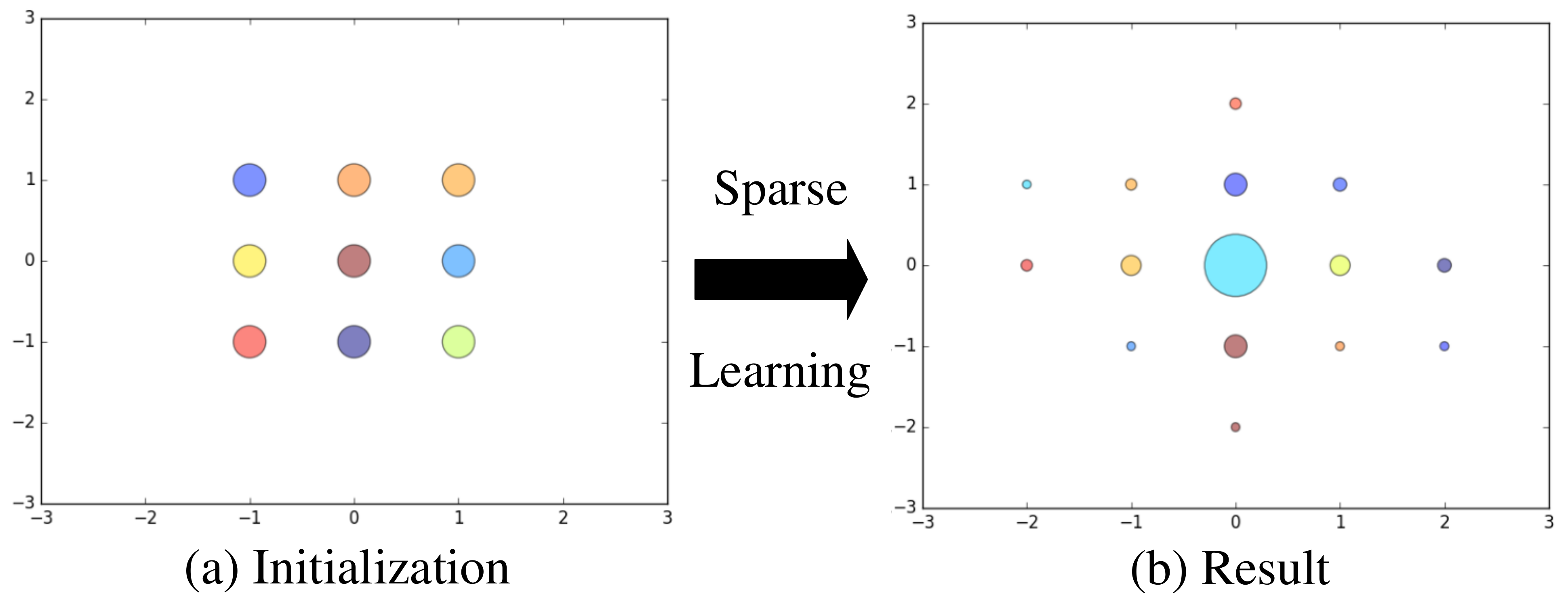}
\end{center}
\caption{The visualization of shift values in the shift layer from block2\_2 of ShiftResNet-20 on CIFAR100. The area of each point denotes the channel number of the feature maps with the same shift pattern. x-axis and y-axis denote the horizontal and vertical displacement, respectively. (Best viewed in color)}
\label{sparsify}
\end{figure}
\begin{table}[tp]
\newcommand{\tabincell}[2]{\begin{tabular}{@{}#1@{}}#2\end{tabular}}
\begin{center}
\begin{tabular}{|c|c|c|c|c|}
\hline
\multirow{2}{*}{Block}&\multicolumn{2}{|c|}{CIFAR10}&\multicolumn{2}{|c|}{CIFAR100}\\
\cline{2-5}
&\tabincell{c}{Unshifts/\\Channels}&\tabincell{c}{Shift\\\!\!Sparsity\!\!}&\tabincell{c}{Unshifts/\\Channels}&\tabincell{c}{Shift\\\!\!Sparsity\!\!}\\
\hline
\!\!block1\_1\!\!&93 / 96&96.9\%&82 / 96&85.4\%\\
\!\!block1\_2\!\!&87 / 96&90.6\%&84 / 96&87.5\%\\
\!\!block1\_3\!\!&94 / 96&97.9\%&92 / 96&95.8\%\\
\hline
\!\!block2\_1\!\!&96 / 96&100\%&96 / 96&100\%\\
\!\!block2\_2\!\!&161 / 192&83.9\%&146 / 192&76.0\%\\
\!\!block2\_3\!\!&181 / 192&94.3\%&164 / 192&85.4\%\\
\hline
\!\!block3\_1\!\!&190 / 192&99.0\%&189 / 192&98.4\%\\
\!\!block3\_2\!\!&331 / 384&86.2\%&316 / 384&82.3\%\\
\!\!block3\_3\!\!&382 / 384&99.5\%&319 / 384&83.1\%\\
\hline
Total&\!\!1615 / 1728\!\!&93.5\%&\!\!1488 / 1728\!\!&86.1\%\\
\hline
\end{tabular}
\end{center}
\caption{The shift sparsity of each layer in ShiftResNet-20 ($\lambda=0.0005$) on CIFAR10 and CIFAR100.}
\label{layer_sparsity}
\end{table}
\begin{table}[tp]
\newcommand{\tabincell}[2]{\begin{tabular}{@{}#1@{}}#2\end{tabular}}
\begin{center}
\begin{tabular}{|c|c|}
\hline
\tabincell{c}{Removed shift layer number\\(ShiftResNet20-SSL, $\lambda$=0)}&\tabincell{c}{Accuracy\\CIFAR10/CIFAR100}\\
\hline
0&91.7\% / 69.2\%\\
\hline
4&91.5\% / 68.2\%\\
\hline
6&91.4\% / 67.0\%\\
\hline
8&89.4\% / 66.0\%\\
\hline
9 (All removed)&81.5\% / 56.7\%\\
\hline
\end{tabular}
\end{center}
\caption{The performance of ShiftResNet-20 on CIFAR10 and CIFAR100 after removing the most unimportant shift layers.}
\label{less_shift_layer}
\end{table}
\begin{table}[tp]
\begin{center}
\begin{tabular}{|c|c|c|c|c|c|}
\hline
Input & Operator & t & n & c & s\\
\hline
$224^2\times 3$ & conv 3$\times$3+BN & - & - & 16 & 2\\
$112^2\times 16$ & IB-SSL & 4 & - & 16 & 1\\
$112^2\times 16$ & IB-Pool 2$\times$2& 5 & - & 32 & 2 \\
$56^2\times 32$&FE-Block&6&3&64&2\\
$28^2\times 64$&FE-Block&6&4&128&2\\
$14^2\times 128$&FE-Block&6&4&128&1\\
$14^2\times 128$&FE-Block&6&4&256&2\\
$7^2\times 256$&FE-Block&6&3&256&1\\
$7^2\times 256$&conv1$\times$1+BN+ReLU &-&-&1380&1\\
$7^2\times 1380$& GAP 7$\times$7&-&-&1380&-\\
$1^2\times 1380$& Dropout 0.2&-&-&1380&-\\
$1^2\times 1380$& conv 1$\times$1&-&-&1000&1\\
\hline
\end{tabular}
\end{center}
\caption{Network configuration. t denotes expansion rate. n means the computational unit number of FE-Block. c denotes the output channels. And s means stride.}
\label{configuration}
\end{table}
\begin{table}[tp]
\newcommand{\tabincell}[2]{\begin{tabular}{@{}#1@{}}#2\end{tabular}}
\begin{center}
\begin{tabular}{|c|c|c|c|}
\hline
Networks&\!\!\!MAdds\!\!\!& \!\!Params\!\!&Top-1\\
\hline
MobileNetV1 0.75x \cite{MobileNets}&325M&2.6M&68.4\%\\
MobileNetV2 1.0x\cite{sandler2018mobilenetv2:}& 300M & 3.4M & 72.0\%\\
\!\!\!ShuffleNetV1 1.5x(g=3)\cite{ShuffleNets} \!\!\!& 292M & 3.4M & 69.0\%\\
ShuffleNetV2 1.5x\cite{shufflenetv2}& 299M & 3.5M & 72.6\%\\
IGCV3-D \cite{igcv3}& 318M & 3.6M & 72.2\%\\
CondenseNet(G=C=8)\cite{condensenet} & 274M & 2.9M & 71.0\%\\
ShiftNet-B \cite{wu2018shift:} & 371M & 1.1M & 61.2\%\\
AS-ResNet-w50 \cite{activeShift}& 404M & 1.96M & 69.9\%\\
FE-Net (ours) 1.0x & 301M & 3.7M & {\bfseries 72.9\%}\\
\hline
MobileNetV1 1.0x\cite{MobileNets}&569M&4.2M&70.6\%\\
MobileNetV2 1.4x\cite{sandler2018mobilenetv2:}& 585M & 6.9M & 74.7\%\\
ShuffleNetV1 2x\cite{ShuffleNets} & 524M & 5.4M & 70.9\%\\
ShuffleNetV2 2x\cite{shufflenetv2}& 591M & 7.4M & 74.9\%\\
IGCV3-D 1.4x\cite{igcv3}& 610M & 7.2M & 74.55\%\\
CondenseNet(G=C=4)\cite{condensenet}& 529M & 4.8M & 73.8\%\\
PNASNet\cite{PNASNET}&588M&5.1M&74.2\%\\
DARTS \cite{DARTS}&595M&4.9M&73.1\%\\
ShiftNet-A \cite{wu2018shift:} & \!\!1400M\!\!& 4.1M & 70.1\%\\
AS-ResNet-w68 \cite{activeShift}& 729M & 3.42M & 72.2\%\\
FE-Net (ours) 1.375x & 563M & 5.9M & {\bfseries 75.0\%}\\
\hline
\end{tabular}
\end{center}
\caption{The performance comparison of several compact neural architectures on ImageNet.}
\label{ImageNetAccuracy}
\end{table}

We take ShiftResNet-20 on CIFAR10 and CIFAR100 with $\lambda$ = 5e-4 for more detailed study. In Tab.\ref{layer_sparsity}, we show the shift sparsity of each layer in detail. In some of the blocks, almost all feature maps stay unshifted which indicates the shift layers in these positions are unimportant. Actually, the sparsity of shift layer can be taken as a metric to measure the importance of these layers. It can decide which shift layer is unimportant and can be removed without accuracy decline. For examples, the shift layer in block2\_1 is the most unimportant while the one in block2\_2 is the most important in ShiftResNet-20. We take the shift layer in block2\_2 for visualization as shown in Fig.\ref{sparsify}. Although the majority of channels stay unshifted, the remaining ones can learn a meaningful shift pattern and provide multiple receptive fields. Actually, a shift layer cooperated with point-wise convolution can take a role of an Inception module. This is also a major advantage of shift layer over conventional convolution layer.

To take a further analysis, we carry out several experiments with ShiftResNet-20 on CIFAR10 and CIFAR100 by removing the most unimportant shift layers according to their sparsity in Tab.\ref{layer_sparsity}. As shown in Tab.\ref{less_shift_layer}, when we progressively remove the most unimportant shift layers, the accuracy only declines a little. Even when we preserve only one shift layer in block2\_2, the accuracy still maintains in a considerable level.

\subsection{Performance on ImageNet}\label{ImageNet_Experiments}

Our redesigned network architecture for ImageNet2012 classification task is described in Tab.\ref{configuration}, which is mainly composed by FE-Block equipped with SSL. We use width multiplier as a hyperparameter to tune the tradeoff between accuracy and computational cost. 

{\bfseries Comparison with other counterparts.} As shown in Tab.\ref{ImageNetAccuracy}, with network architecture improvement, our results surpass ShiftNet and AS-ResNet by a large margin. What's more, before our work, the best performance in terms of FLOPs/accuracy is always dominated by the networks built by depthwise separable convolution in the past few years. We are the first one to build a compact neural network without using depthwise separable convolution which can achieve superior results to other counterparts constructed by depthwise separable convolution. As shown in Tab.\ref{ImageNetAccuracy}, our network surpasses MobileNet series networks and ShuffleNet series networks, as well as the networks automatically searched by NAS technique \cite{PNASNET,DARTS}, indicating that SSL can be taken as an alternative choice over depthwise separable convolution. This can provide a new basic component for NAS and inspire more exploration in this direction.

\begin{figure}[tp]
\begin{center}
\includegraphics[scale=0.35]{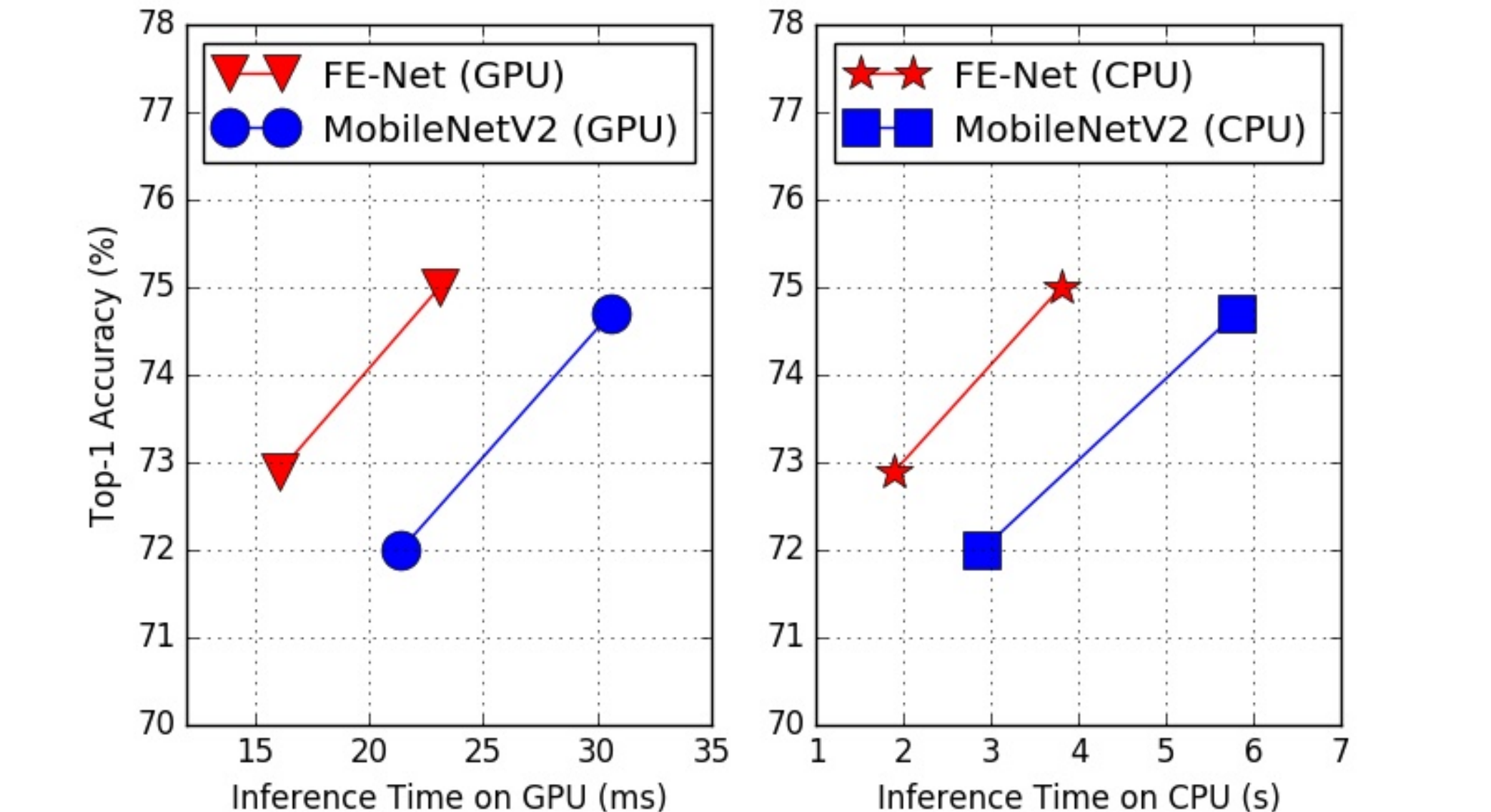}
\end{center}
\caption{Practical runtime comparison with MobileNetV2.}
\label{runtime}
\end{figure}

As for practical runtime, we mainly compare our network with MobileNetV2, which is the most representative compact network constructed by depthwise separable convolution. As illustrated in Fig.\ref{runtime}, our networks achieve higher accuracy with significantly faster inference time on GPU and CPU, which proves that SSL is a more friendly basic component for practical application scenarios.

{\bfseries An ablation study of FE-Net.} We also train the FE-Nets equipped with depthwise separable convolution (DW) on ImageNet so as to decompose the benifit of SSL from the improved network design. As shown in Tab.\ref{FE-Net}, the gap of accuracy between SSL and DW based FE-Net is small while their practical runtime is significant larger, which further validates the superiority of SSL and FE-Net.

\begin{figure}[tp]
\begin{center}
\includegraphics[scale=0.21]{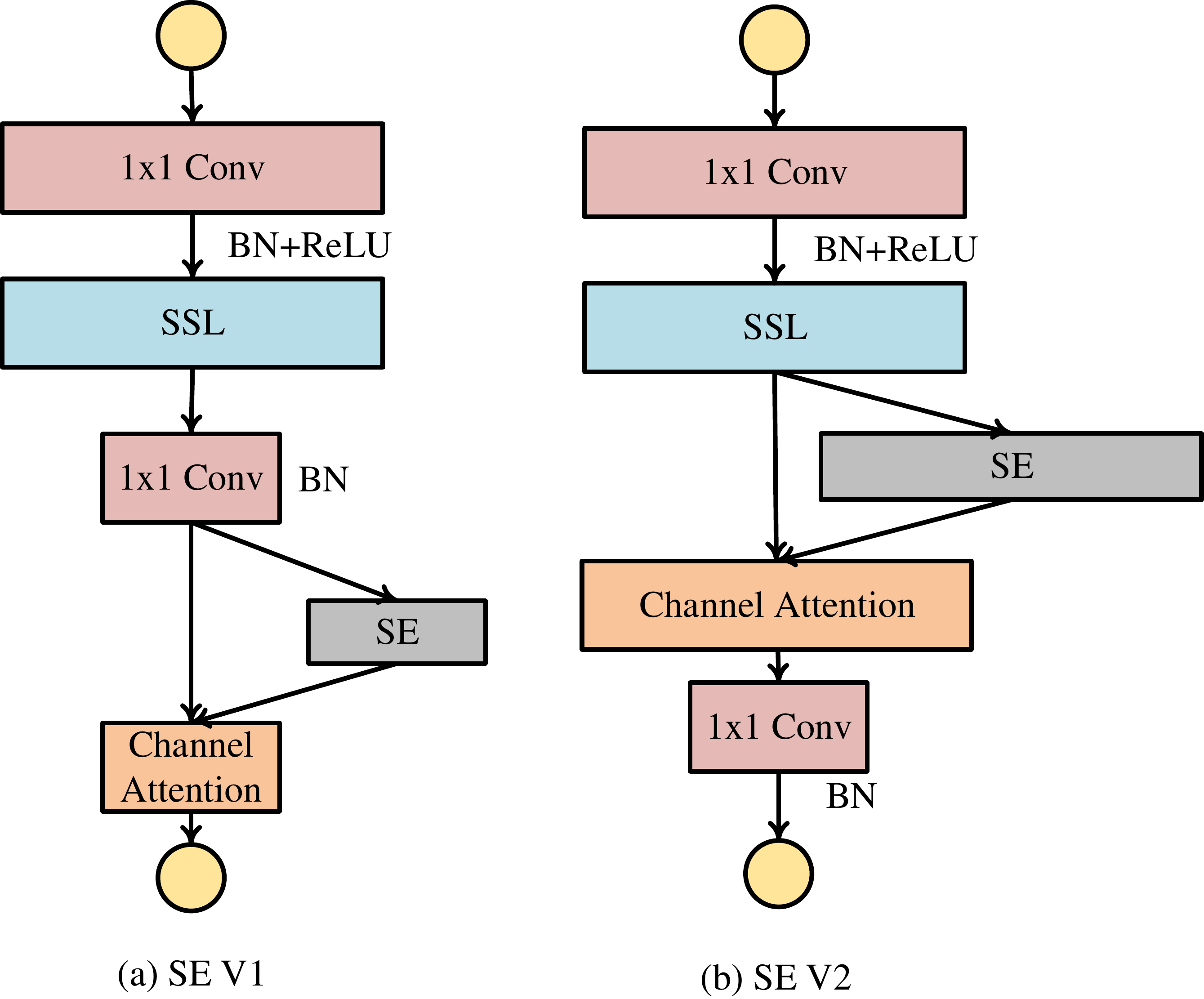}
\end{center}
\caption{Two different placement manners of SE module.}
\label{SE_position}
\end{figure}

{\bfseries Compatibility with other methods.} Our network can also be combined with other methods for further performance exploration. For instance, our network can be equipped with SE module (\emph{Squeeze-and-Excitation} \cite{SE}) for channel attention. However, we find it matters to place SE module in different position of basic block. Here we only discuss the position of SE module in inverted bottleneck. As illustrated in Fig.\ref{SE_position}, there are two different placement manners. The first manner is the conventional one, which places SE module in the output position of the basic block. However, as for inverted bottleneck, the most redundant information exists in the expansion part. Since SE module is used for channel attention, it is more reasonable to place SE module in the expansion part of inverted bottleneck as shown in Fig.\ref{SE_position}(b). The results in Tab.\ref{SE_result} empirically validates this idea. Moreover, we note that the shift sparsity increases a lot after equipping SE module as shown in Tab.\ref{sparsity_SE_result}. Through channel-wise feature recalibration, it imposes more unshifted feature maps. This can reflect the power of channel attention brought by SE module from another perspective.

\begin{table}[htp]
\begin{center}
\begin{tabular}{|c|c|c|c|c|}
\hline
Networks & Top1&Top5&GPU&CPU\\
\hline
FE-Net & 72.9\% & 91.2\%&16.1ms&1.9s\\
FE-Net (DW)& 73.2\% & 91.4\%&21.8ms&2.7s\\
\hline
FE-Net 1.375x& 75.0\% & 92.4\%&23.1ms&3.8s\\
\!\!\!FE-Net 1.375x (DW)\!\!\!& 75.2\% & 92.8\%&30.4ms&5.3s\\
\hline
\end{tabular}
\end{center}
\caption{An ablation study of FE-Net with shift operation (SSL) vs. depthwise convolution (DW) on ImageNet (batchsize 32).}
\label{FE-Net}
\end{table}
\begin{table}[tp]
\newcommand{\tabincell}[2]{\begin{tabular}{@{}#1@{}}#2\end{tabular}}
\begin{center}
\begin{tabular}{|c|c|c|c|}
\hline
Networks&\!\!MAdds\!\!&\!\!Params\!\!&Top1\\
\hline
\!\!MobileNetV1 1.0x + SE \cite{SE}\!\!& 572M & 4.7M & 74.7\%\\
\hline
\!\!ShuffleNetV2 2X + SE \cite{shufflenetv2}\!\!& 597M & - & 75.4\%\\
\hline
\!\!FE-Net 1.375x + SE V1\!\!& 564M & 6.1M & 75.6\%\\
\hline
\!\!FE-Net 1.375x + SE V2\!\!& 566M & 8.2M & \!\!{\bfseries 76.5\%}\!\!\\
\hline
\end{tabular}
\end{center}
\caption{The performance comparison of several compact neural architectures equipped with SE modules on ImageNet.}
\label{SE_result}
\end{table}
\begin{table}[tp]
\newcommand{\tabincell}[2]{\begin{tabular}{@{}#1@{}}#2\end{tabular}}
\begin{center}
\begin{tabular}{|c|c|c|}
\hline
Networks & Top1 & Shift Sparsity\\
\hline
FE-Net 1.0x & 72.9\%& 60.0\%\\
\hline
FE-Net 1.375x& 75.0\%& 69.5\%\\
\hline
FE-Net 1.375x + SE V1 & 75.6\% & 77.7\%\\
\hline
FE-Net 1.375x + SE V2 &76.5\% & 80.2\%\\
\hline
\end{tabular}
\end{center}
\caption{The shift sparsity of FE-Net with different accuracy.}
\label{sparsity_SE_result}
\end{table}

\section{Conclusions}
In this paper, we mainly study the feasibility of SSL to build a compact and accurate neural network. Extensive experiments prove that only a few shift operations are sufficient for spatial information communication. We also show that SSL can be taken as an efficient alternative over depthwise separable convolution. A well-designed network equipped with SSL can surpass other counterparts equipped with depthwise separable convolution in terms of accuracy, FLOPs and practical inference time. Our work will inspire more exploration for network design and searching.

{\small
\bibliographystyle{ieee}
\bibliography{SSL}
}

\end{document}